\documentclass[letterpaper]{article}

\usepackage{cite, graphicx, amsmath, amsfonts, amssymb, array, latexsym, url, algorithm, algorithmic}
\usepackage{multirow, booktabs}

\newcommand{\xx}{{\bf x}}
\newcommand{\zz}{{\bf z}}
\newcommand{\ls}{\varphi(\xx)}

\begin{document}

\title{Information tracking approach to segmentation of ultrasound imagery of prostate}

\author{Robert~Sheng~Xu,~Oleg~Michailovich, and~Magdy~Salama
\thanks{This research was supported by a Discovery grant from NSERC -- The Natural Sciences and Engineering Research Council of Canada. Information on various NSERC activities and programs can be obtained from {\tt http://www.nserc.ca}.}%
\thanks{R. Xu, O. Michailovich, and M. Salama are with the School of Electrical and Computer Engineering, University of Waterloo, Canada N2L 3G1 (phone: 519-888-4567; e-mails: {rsxu, olegm, M.Salama}@uwaterloo.ca).}}

\maketitle

\begin{abstract}
The size and geometry of the prostate are known to be pivotal quantities used by clinicians to assess the condition of the gland during prostate cancer screening. As an alternative to palpation, an increasing number of methods for estimation of the above-mentioned quantities are based on using imagery data of prostate. The necessity to process large volumes of such data creates a need for automatic segmentation tools which would allow the estimation to be carried out with maximum accuracy and   efficiency. In particular, the use of transrectal ultrasound (TRUS) imaging in prostate cancer screening seems to be becoming a standard clinical practice due to the high benefit-to-cost ratio of this imaging modality. Unfortunately, the segmentation of TRUS images is still hampered by relatively low contrast and reduced SNR of the images, thereby requiring the segmentation algorithms to incorporate prior knowledge about the geometry of the gland. In this paper, a novel approach to the problem of segmenting the TRUS images is described. The proposed approach is based on the concept of distribution tracking, which provides a unified framework for modeling and fusing image-related and morphological features of the prostate. Moreover, the same framework allows the segmentation to be regularized via using a new type of ``weak" shape priors, which minimally bias the estimation procedure, while rendering the latter stable and robust. The value of the proposed methodology is demonstrated in a series of both {\it in silico} and {\it in vivo} experiments.
\end{abstract}


\newpage
\section{Introduction}\label{Intro}
Ultrasound imaging has long become an integral part of modern health care due to its superiority over many alternative imaging modalities in terms of its cost-to-benefit ratio. In particular, the properties of ultrasound imaging being practically harmless, mobile, readily accessible, and cost efficient has made it the modality of choice in many clinical settings. Unfortunately, the above advantages have always been counterbalanced by the relatively low quality of ultrasound images as compared to X-ray CT or MRI scans. As a result, there has always been a need for image processing tools which, when applied to ultrasound images, could produce valuable outcomes under the conditions of reduced resolution and contrast~\cite{Suetens09}.

According to recent global cancer statistics, prostate cancer constitutes the fifth most common type of cancer in the world and the second most common in men~\cite{Parkin02}. The medical treatment cost of prostate cancer runs into millions of dollars annually, while the emotional cost to the patients and the members of their families is incalculable. This is why early diagnosis of prostate cancer through systematic screening has become the main tool of prevention of the disease. Even though the current practice of screening is still based on palpation, it is believed that more reliable diagnostic outcome is possible via using minimally invasive procedures, such as {\it transrectal ultrasound imaging} (TRUS)\cite{Bluth07}. Moreover, since the size and shape of the prostate are among the main quantities used by clinicians to identify the presence of abnormal developments within the gland~\cite{UPMC}, automatic segmentation of ultrasound prostate images is currently recognized as a pivotal component of the TRUS-based diagnosis of prostate cancer~\cite{Pathak00, Gong04, Yu04, Shen06, Yang06}.

In the case of TRUS imaging, the existing methods of prostate segmentation encompass a number of different strategies. Some basic solutions to the problem are based on utilizing edge-related features pertaining to the prostate boundary~\cite{Ladak00, Pathak00, Yu04}. For example, the method of~\cite{Ladak00} prompts the user to manually input a set of initial points on the boundary of the gland. These points are used to initialize a discrete dynamic contour~\cite{Lobregt95}, which is subsequently driven towards strong local edges. Alternatively, the approach of~\cite{Pathak00} detects the prostate boundary using the Canny edge detector applied to the TRUS images enhanced by means of anisotropic diffusion filtering. The resulting edge maps are then superimposed over the original images as a visual guide to subsequent manual delineation. Note that, even though this method has not been devised as a fully automatic procedure, it could still be useful to reduce both intra- and inter-observer variances of prostate delineation. To improve the robustness of prostate segmentation in the case of poorly observable edges, a two step segmentation procedure was proposed in~\cite{Yu04}. At the first stage of this method, a coarse segmentation is obtained based on texture-related features of the TRUS images, followed by forcing the segmentation boundary to converge to local image edges during the second stage. Unfortunately, this method (as well as all the methods mentioned before) has a drawback of being prone to the errors caused by the existence of spurious (noise-induced) edges and shadowing artifacts.

Another classical approach to the problem of prostate segmentation takes advantage of the tools of
supervised~\cite{Prater92, Zaim07} and unsupervised~\cite{Richard96} machine learning. Specifically, the method of~\cite{Prater92} employs a feed-forward neural network to carry out the segmentation, which attempts to capture the visual cues used by experts to classify image pixels as either inside or outside of the prostate region. In the recent approach of~\cite{Zaim07}, the image features used by a support vector machine classifier are defined via projecting the TRUS images onto a basis of multi-wavelets. On the side of unsupervised learning, a probabilistic clustering procedure detailed in~\cite{Richard96} segments the prostate region based on the features obtained using a specially designed filter bank~\cite{Laws80}. Despite the strong theoretical foundation of the classification-based segmentation algorithms, they remain prone to over-segmentation because of their disregard for the morphological features of the prostate gland.

Recent efforts on prostate segmentation have been extended to use the information provided by {\em shape priors} to regularize the process of segmentation~\cite{Gong04, Saroul08, Shen03, Shen06}. Thus, for example, the method in~\cite{Gong04} models the prostate boundary as a deformable super-ellipse. Subsequently, a maximum-a-posteriori estimation framework is used to find a contour (i.e. a boundary of the prostate) that closely matches the prior shape model on one hand, and coincides with strong image gradients on the other. The same super-ellipse model is employed in~\cite{Saroul08} where texture-related features are used instead of edge-related ones. An obvious limitation of the above approaches stems from the restrictive nature of the prior model used, which can only approximate symmetric shapes. More general assumptions regarding the shape of prostate are used in~\cite{Shen03}, which uses a Gabor filter bank to describe the prostate boundary. Subsequently, a hierarchical shape deformation procedure moves an initial segmentation towards the location of the true boundaries of the gland. The implementation of this method, however, is hindered by a number of initialization procedures such as a normalization of the shape of the ultrasound transducer in use, and a geometric transformation of the TRUS images to align them with a prior shape model. Moreover, the use of a greedy energy-minimization algorithm to search for an optimal prostate boundary leads to relatively large computation times, which makes the above method impractical to apply to large data sets. 

Among other methods for prostate segmentation, one can mention the semi-automatic procedure of~\cite{Zouqi08} which is based on the method of graph cuts, and the approach in~\cite{Abolmaesumi04} which takes advantage of a Kalman state estimator~\cite{Kirburajan98} to track the prostate boundary starting from a seed location.   

In the current paper, a different approach to the problem of segmentation of TRUS images is introduced. As opposed to the previous algorithms described above, the proposed method is exceptional for it concurrently fulfills a number of essential objectives, {\it viz.}
\begin{enumerate}
\item {\it Generality:} The algorithm fuses information from a range of different sources, {\it viz.} edge-related, texture-related, and shape-related features are all used to estimate an optimal boundary of the prostate gland.
\item {\it Adaptability:} The method allows using any number of arbitrary texture-related features of the  TRUS images such as their intensity, local polynomial moments, multiresolution coefficients, local entropies,  etc. Moreover, the method can also accommodate the use of any number of shape descriptors (subject to some transformation-invariance requirements to be discussed below).
\item {\it Consistency:} The method takes advantage of the concept of distribution tracking~\cite{Freedman04}, which allows applying the same methodology to handle both photometric and morphological features of the prostate, thereby providing a unified framework for the segmentation problem at hand.  
\end{enumerate}

Finally, it is important to note that integrating shape prior models into probabilistic image segmentation has never been a trivial task, with a spectrum of different approaches proposed to this end~\cite{Paragios06}. In most of the cases, the antecedent knowledge on plausible configurations of an optimal shape is ``encoded" in the form of prior probabilities which can describe the distribution of, e.g., the control points~\cite{Blake92} or level-set functions~\cite{Tsai03} related to the shape representation. Such methods have proven particularly useful in the cases when the objects to be segmented appear to be partially observable due to, e.g., occlusions. Under these circumstances, the use of elaborate shape priors seems to be well justified, even when a sizable bias is introduced into the estimated shapes. For the case at hand, however, using such priors would {\em not} be a proper choice, since the prostates are rarely (if ever) occluded in TRUS imaging. In this situation, the prior information should be restrictive just enough to regularize the process of segmentation, which should still be dominated by the information contained in the acquired images.

To achieve the above goal, a new type of ``weak" shape priors is introduced in this paper. The prior information is encapsulated in the form of a probability density function (pdf) of an intrinsic geometric parameter (or a set thereof) of the true prostate boundary. Subsequently, the segmentation boundary is forced to evolve into a configuration whose empirical pdf of the same geometric parameter(s) closely matches the model pdf as assessed by the Bhattacharyya distance measure. In this sense, the present paper extends the approaches of~\cite{Freedman04, Oleg07} to tracking {\em both} morphological and photometric features of the object of interest. Moreover, it is proven experimentally that the proposed methodology can be effectively used for robust and accurate segmentation of TRUS images under the conditions of low image contrast and poorly observable prostate boundaries. 

The rest of the paper is organized as follows. In Section II, a method for prostate image segmentation via tracking of both texture-related and morphological features is detailed. A number of essential technical considerations are provided in Section III, while Section IV presents a series of both {\it in silico} and {\it in vivo} experiments. Finally, Section V summarizes the paper with a discussion, conclusions, and an outline of our future research.

\section{Tracking of distributions}
\subsection{Tracking texture-related features}\label{TrackFeat}
Let $u(\xx)$ be an ultrasound image which we consider to be a scalar-valued function defined over a subset $\Omega$ of $\mathbb{R}^2$. To render our discussion general, let $\mathcal{M}$ be a map which transforms $u(\xx)$ into its corresponding {\em feature image} $I(\xx)=\mathcal{M}[u(\xx)]$, which can be viewed as a vector-valued image (i.e. $I(\xx): \Omega \rightarrow \mathbb{R}^d$ ) with its components being the features pertaining to $u(\xx)$. Thus, for each $\xx \in \Omega$, $I(\xx)$ may consist of the values of the original image $u(\xx)$, its multiresolution version~\cite{Unser95}, some phase-related and similarity measures~\cite{Noble06}, etc.  

Let $\Omega_t$ (with the subscript $t$ standing for ``target") be a subset of $\Omega$ over which the object of interest (i.e. prostate) is supported. The segmentation algorithm proposed in this paper is based on the following two assumptions. First, for each $\xx \in \Omega_t$, the $d$ components of $I(\xx)$ are assumed to be independent random variables\footnote{Note that the assumption of independence has been mainly introduced to render the algorithm computationally feasible, and thus a mitigation of this assumption should be addressed in future research.}. Second, it is assumed that one is provided with reliable estimates of the {\em probability density functions} (pdf) of the image features pertaining to the prostate region $\Omega_t$. Note that, even though the second assumption could seem impractical, it is actually easy to attain due to the availability of vast depositories of manually segmented prostate images. These images can readily be used to extract the values of the features of interest within the boundaries of manually delineated prostates. These values in turn can be used to estimate their corresponding pdf's by means of any standard procedure~\cite{Silverman86}.

Let $P_t(\zz)$, where $\zz \in \mathbb{R}^d$, be the joint pdf of the image features associated with the prostate region $\Omega_t$. Due to the assumption of statistical independence of the image features, $P_t(\zz)$ can be factorized as $P_t(\zz) = \prod_{k=1}^d p_t(z_k)$, with $p_t(z_k)$ being the pdf of the $k$-th image feature $z_k \in \mathbb{R}$. Now, let $\Omega_{in} \subseteq \Omega$ be an arbitrary subset of $\Omega$ (which is not equal to $\Omega_t$, in general). Then, for a given $I(\xx)$ one can define the {\em empirical} probability density of $z_k$ observed over $\Omega_{in}$ as  
\begin{equation}\label{E1}
p(z_k \mid \Omega_{in}) = \frac{\int_{\Omega_{in}} K(z_k - I_k(\xx)) \, d\xx}{\int_{\Omega_{in}} d\xx},
\end{equation}
where $I_k(\xx)$ is the $k$-th component of $I(\xx)$, and $K$ is a positive-valued {\it kernel} function satisfying $\int K(x) dx = 1$. Note that the expression in (\ref{E1}) is nothing else but a kernel-based, nonparametric estimate of the pdf of $z_k$~\cite{Silverman86, Simonoff96}. By virtue of the statistical independence of $z_k$, the joint (empirical) pdf of $\zz$ observed over $\Omega_{in}$ is given by $P(\zz \mid \Omega_{in}) = \prod_{k=1}^d p(z_k \mid \Omega_{in})$. Consequently, one can quantify the degree of similarity between the {\it target} $P_t(\zz)$ and the {\it empirical} $P(\zz \mid \Omega_{in})$ densities by means of the Bhattacharyya coefficient $B$ as given by~\cite{Bhattacharyya43, Kailath67, Goudail04}
\begin{equation}\label{E2}
B(\Omega_{in}) = \int_{\mathbb{R}^d} \sqrt{P_t(\zz) P(\zz \mid \Omega_{in})} \, d\zz.  
\end{equation}
Alternatively, using the assumption of mutual independence of $z_k$, one can rewrite (\ref{E2}) as
\begin{equation}\label{E3}
B(\Omega_{in})=\prod_{k=1}^d B_k(\Omega_{in}),
\end{equation}
where
\begin{equation}\label{E4}
B_k(\Omega_{in})=\int \sqrt{p_t(z_k) \, p(z_k \mid \Omega_{in})} \, dz_k.
\end{equation}

Viewed as a function of $\Omega_{in}$, $B(\Omega_{in})$ takes its values in the interval $[0, \, 1]$, with the maximum attained when $p_t(z_k) = p(z_k \mid \Omega_{in}), \forall k=1,2,\ldots,d$. In particular,  $B(\Omega_{in})$ is maximized when the regions $\Omega_{in}$ and $\Omega_t$ coincide with each other (meaning $\Omega_{in} \equiv \Omega_t$). Consequently, the image $u(\xx)$ can be segmented via approximating $\Omega_t$ by $\Omega_{in}$ which locally maximizes $B(\Omega_{in})$~\cite{Freedman04}.
 
\subsection{Level-set segmentation paradigm}\label{LvlSeg}
It goes without saying that maximizing $B(\Omega_{in})$ as a function of $\Omega_{in}$ would result in a combinatorial optimization of little practical interest. To find a useful segmentation in a consistent manner, a different formulation of the problem is needed. In particular, in this paper, $\Omega_{in}$ is defined {\it implicitly} as a subset of $\Omega$, over which a {\it level-set function} $\varphi(\xx): \Omega \rightarrow \mathbb{R}$ attains non-positive values, {\it viz.}
\begin{equation}
\Omega_{in} = \left\{ \xx \in \Omega \mid \varphi(\xx) \leq 0 \right\}.
\end{equation}
Consequently, using the standard definition of the Heaviside function as $\mathcal{H}(x) = (x)_+$, the Bhattacharyya coefficient in (\ref{E3}) can be redefined as a function of $\varphi(\xx)$ as given by
\begin{equation}\label{E5}
B(\varphi(\xx)) = \prod_{k=1}^d B_k(\varphi(\xx)) = \prod_{k=1}^d \int \sqrt{p_t(z_k) \, p(z_k \mid \varphi(\xx))} \, dz_k,
\end{equation}
where
\begin{equation}\label{E6}
p(z_k \mid \varphi(\xx)) = \frac{\int_{\Omega} K(z_k - I_k(\xx)) \, \mathcal{H}(-\varphi(\xx))\, d\xx}{\int_{\Omega}  \mathcal{H}(-\varphi(\xx)) \, d\xx}.
\end{equation}
Consequently, the problem of segmentation of $u(\xx)$ amounts to finding an optimal $\varphi(\xx)$ which maximizes $B(\varphi(\xx))$.

A (local) maximizer of $B(\varphi(\xx))$ can be computed via the standard {\it steepest ascent} procedure which has the form of a gradient flow defined as
\begin{equation}\label{E7}
\varphi_\tau(\xx, \tau) \triangleq \frac{\partial \varphi(\xx, \tau)}{\partial \tau} = \frac{\delta B(\varphi)}{\delta \varphi},
\end{equation} 	
with $\tau$ being an artificial time parameter (so that $\varphi(\xx,0) = \varphi_0(\xx)$ is an initialization of the level-set function), and $\delta B(\varphi) \slash \delta \varphi$ standing for the first variation of $B(\varphi(\xx))$ computed with respect to $\varphi(\xx)$. Appendix A provides necessary technical details on the derivation of $\delta B(\varphi) \slash \delta \varphi$, and shows that (\ref{E7}) can be rewritten as given by
\begin{equation}
\varphi_\tau(\xx, \tau) = \delta(\varphi(\xx, \tau)) \, V_B(\xx, \tau),
\end{equation} 
where $\delta(\cdot)$ stands for the Dirac delta function, and $ \delta(\varphi(\cdot)) V_B(\cdot): \Omega \rightarrow \mathbb{R}$ is a velocity field that governs the evolution of the level-set $\ls$. (In the developments below, we suppress the explicit dependency of both level-set functions and velocity fields on $\tau$, until it is needed to help clarify the derivations).

It should be noted that the form of the {\em gradient flow} in (\ref{E7}) implies that the latter depends on the observed data $I(\xx)$ alone, while disregarding some plausible properties of the optimal solution. As a result, maximizing (\ref{E5}) could be too sensitive to measurement noises and/or errors in the data. This sensitivity, however, can be alleviated via regularizing the maximization of $B(\ls)$ using the framework of {\em geodesic active contours}~\cite{Caselles97, Yezzi97}. In this case, the optimal $\varphi(\xx)$ is found as
\begin{equation}\label{E8}
\varphi^\star(\xx) = \arg\sup_{\varphi (x)} \left\{ \alpha \, B(\varphi(\xx)) - \int_{\Omega} g(\xx)\,  \| \nabla \mathcal{H}(\varphi(\xx))\| \, d\xx \right\},
\end{equation}
where $\nabla$ denotes the operator of gradient, $\|\cdot\|$ is the Euclidean norm, $\alpha >0$ is a regularization constant\footnote{In general, $\alpha$ can be a user defined parameter. In the current paper, its value has been set to be equal to 0.5.}, and $g: \Omega \rightarrow \mathbb{R}^+$ is an edge-detector function (to be defined below). As a result, the gradient flow associated with the optimization problem in (\ref{E8}) can be shown to be equal to
\begin{equation}\label{E9}
\varphi_\tau(\xx) = \delta(\varphi(\xx)) \left(\alpha \, V_B(\xx) + {\rm div} \left( g(\xx) \frac{\nabla \varphi(\xx)}{\| \nabla \varphi(\xx) \|} \right)\right),
\end{equation}

It is worth pointing out that the gradient flow in (\ref{E9}) is designed to converge to an optimal level-set function, whose related $\Omega_{in}$ has its boundary aligned with the strong edges of $u(\xx)$, while the empirical distributions of the image features observed over $\Omega_{in}$ are ``aligned" with the model distributions $p_t(z_k)$ in the sense of maximizing $B(\ls)$. 

Finally, it should be noted that the gradient flow (\ref{E9}) would change the level-set function $\varphi(\xx)$ over a set of zero measure, if the formal delta function $\delta(\cdot)$ was used in computations. To overcome this ``technical'' difficulty, it is common to extend the numerical support of the level-set evolution via replacing $\delta(\cdot)$ by its smoother version $\delta_\epsilon(\cdot)$ that could be defined as given by, e.g.,~\cite{ChanVese01}:
\begin{equation}\label{E10}
\delta_{\epsilon}(x) =
\begin{cases}
\frac{1}{2 \epsilon} \left(1 + \cos(\frac{\pi x}{\epsilon})\right), & |x| \leq \epsilon \\
0, &{\rm otherwise}
\end{cases}
\end{equation}
Note that the support of $\delta_{\epsilon}$ is finite and controlled though the user-defined parameter $\epsilon$. Throughout the experimental study of this paper, $\epsilon$ is set to be equal to 2. 

\subsection{A Motivating Example}
In practical computations, the boundary $\Gamma_{in}$ of $\Omega_{in}$ is used akin to a decision boundary that separates the object of interest from its surrounding. This boundary -- conventionally referred to as an {\em active contour} -- is defined by the {zero} level set of $\varphi(\xx)$, i.e. $\Gamma_{in} \triangleq \{\xx \in \Omega \mid \varphi(\xx)=0\}$. In the present case, the active contour belongs to the family of geodesic active contours, which are particularly useful in the situations when objects of interest have well-defined edges. Unfortunately, in ultrasound imaging, edges can hardly be considered as reliable features. Due to shadowing artifacts, diffraction, as well as the destructive nature of speckle noise, the ultrasound representation of many anatomical structures (including prostates) happens to lack continuously defined boundaries. Consequently, using the segmentation tools dependent on an edge-detection procedure is prone to converge to erroneous results in the case of ultrasound imaging.

\begin{figure}[top]
\centering
\includegraphics[width=4.5in]{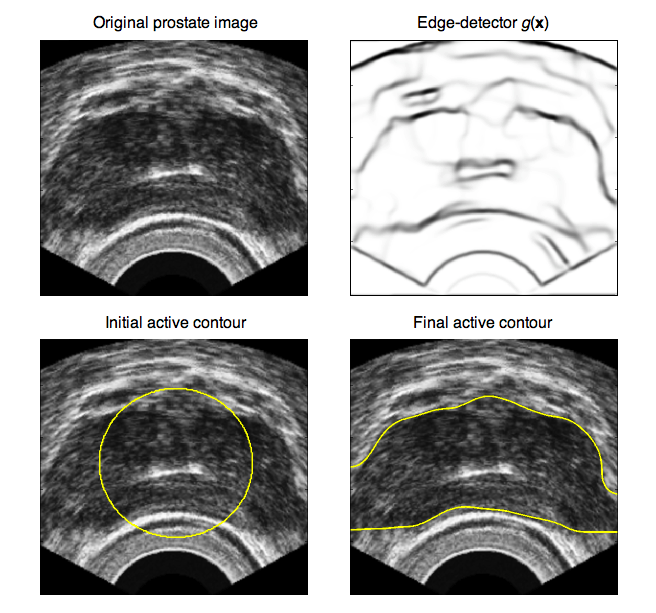} 
\caption{Segmentation of prostate image without shape priors: (A) Original image; (B) Edge-detector function $g(\xx)$; (C) Initial segmentation; (D) Final segmentation.}
\label{F1}
\end{figure}

The above problem is exemplified in Fig.~\ref{F1} which shows an original prostate image (Subplot A), its corresponding edge-detector function $g(\xx)$ (Subplot B), an initial segmentation (Subplot C), and a final segmentation (Subplot D) as obtained by (\ref{E9}). Note that in this experiment, the edge-detector function $g(\xx)$ was defined as\footnote{In general, an edge-detector function is defined to be strictly positive over the homogeneous regions of an image, while vanishing in vicinity of the strong gradients of the image.}
\begin{equation}\label{E11}
g(\xx) = \frac{1}{1 + \lambda \, \| \nabla \tilde{u}(\xx) \|^2},
\end{equation} 
where $\tilde{u}(\xx)$ stands for a de-speckled version of $u(\xx)$ computed by the SRAD filter~\cite{Yu02}, and $\lambda$ is a positive constant (set to be equal to 3 in this example). Moreover, the Bhattacharyya functional in (\ref{E5}) was defined using $I_1(\xx) = u(\xx)$ and $I_2(\xx) = \tilde{u}(\xx)$. One can see that the incompleteness of the prostate boundary along with the shadowing artifacts have caused the active contour to ``leak out" at the sides of the prostate region. To alleviate this deficiency, additional information on the expected geometry of prostate boundary should be incorporated into the segmentation procedure, as explained next.

\subsection{Tracking of Shape Features using Weak Priors}
To overcome the limitation presented in the example of Fig.~\ref{F1} and to guarantee the convergence of the active contour to a useful segmentation, the gradient flow in (\ref{E9}) should be further constrained. One way to achieve this would be to use {\em shape priors} for $\Gamma_{in}$, which would restrict the active contour to converge within a predefined space of expected configurations~\cite{Cootes95, Leventon00, Faugeras00, Paragios02, Tsai03, Cremers04, Bresson06, Paragios06}. A conceptually new solution to the problem of shape priors is proposed in this paper. In particular, we propose to {\em track} the correct shape configuration in the manner similar to that used for tracking of the feature distributions. As will be shown below, the proposed approach is simple to implement as it requires neither registration of training images nor using additional computational structures apart from what has already been used before in the paper. An even more significant property of the proposed prior model is its {\em minimally restrictive} nature. The latter implies that the force produced by the model will be dominated by data-related forces, while remaining sufficiently ``strong" to prevent the segmentation from diverging at the points of poorly observed sections of the prostate boundary.  

To construct such a ``weak" prior model, we suggest to select a single parameter of an expected prostate boundary $\Gamma_t$, whose empirical pdf can be learned based on the same set of manually segmented images which have already been used to compute $P_t(\zz)$. Subsequently, the gradient flow (\ref{E9}) could be redefined in such a way that the empirical pdf of the same parameter of the active contour is forced to match, as close as possible, the learned pdf in terms of the Bhattacharyya coefficient (\ref{E2}).
 
In the current study, we use the {\em curvature} of active contours as the shape-tracking parameter. It is worth noting that the curvature has been chosen for its property of being invariant under the group of Euclidean transformations. Since the configuration of TRUS transducers suggests that different images of the same prostate are likely to be similar up to a rotation and/or a shift, the Euclidean invariance of the curvature allows one to forgo image preprocessing via alignment and registration. However, it should be noted that the proposed approach is not limited to work for the curvature alone, and thus other geometric descriptors of the curves could be used, should one require invariance under different geometric transformations~\cite{Sapiro93, Bruckstein97}.

Given a set of manually delineated images, the fast marching method~\cite{Sethian99} can be used to compute the level-set functions as the signed distance functions of the segmentation boundaries. Subsequently, each of the resulting level set functions $\varphi(\xx)$ can be used to compute its associated curvature according to
\begin{equation}\label{E12}
\kappa(\xx) = - {\rm div} \left\{ \frac{\nabla \varphi(\xx) }{\| \nabla \varphi(\xx) \|} \right\},
\end{equation}
whose (empirical) pdf $C(\xi \mid \ls)$ can in turn be defined as 
\begin{equation}\label{E13}
C(\xi \mid \ls) = \frac{\int_\Omega \delta_\epsilon(\ls) \, K(\xi - \kappa(\xx)) \, d\xx}{\int_\Omega \delta_\epsilon(\ls) \, d\xx},
\end{equation}
where $\kappa(\xx)$ and $\delta_\epsilon(\cdot)$ are given by (\ref{E12}) and (\ref{E10}), respectively.  Note that the calculation in (\ref{E12}) provides the curvature values corresponding to {\em all} level sets of $\ls$. Therefore, $\delta_\epsilon(\ls)$ in (\ref{E13}) makes the resulting pdf depend on the values of $\kappa(\xx)$ in immediate vicinity of the zero level set of $\ls$. Finally, the target curvature distribution $C_t(\xi)$ is obtained via averaging the curvature pdf's  $C(\xi \mid \ls)$ over all available training images.

The same equation (\ref{E13}) can be used to estimate the curvature pdf $C(\xi \mid \varphi(\xx))$ which corresponds to the level-set function $\varphi(\xx)$ used for segmentation. In this case, one can measure the similarity between the target and evolving shapes in terms of the Bhattacharyya coefficient given as
\begin{equation}\label{E14}
B_\kappa(\ls) = \int \sqrt{C_t(\xi) \, C(\xi \mid \ls)} \, d\xi.  
\end{equation} 
Consequently, the shape priors (as defined by $C_t(\xi)$) can be incorporated into the segmentation process via requiring the optimal level-set function $\varphi^\ast(\xx)$ to maximize both $B(\ls)$ in (\ref{E5}) and $B_\kappa(\ls)$ in (\ref{E14}). In this case, the resulting optimization problem takes the form of
\begin{equation}\label{E15}
\varphi^\ast(\xx) = \arg\sup_{\varphi (x)} \left\{ \alpha \, B(\ls) + \beta \, B_\kappa(\ls) - \int_{\Omega} g(\xx)\,  \| \nabla \mathcal{H}(\ls)\| \, d\xx \right\},
\end{equation}
where $\beta > 0$ is a shape-regularization parameter (set to be equal to 2.5 in the experiments reported in this paper). As a result, the optimal solution can now be found as a stationary point of the gradient flow defined as
\begin{equation}\label{E16}
\frac{\partial \varphi(\xx)}{\partial \tau} = \delta_\epsilon(\ls) \left( \alpha \, V_B(\xx) + {\rm div} \left( g(\xx) \frac{\nabla \varphi(\xx)}{\| \nabla \varphi(\xx) \|} \right) \right) + \beta \, V_C(\xx),
\end{equation}
where the velocity $V_C(\xx): \Omega \rightarrow \mathbb{R}$ is related to the maximization of $B_\kappa(\ls)$. The derivation of $V_C(\xx)$ is provided in Appendix B.  

Concluding this section we note that, in this paper, the gradient flow in (\ref{E16}) was computed using the implicit discretization scheme based on the method of additive operator splitting (AOS) as detailed in~\cite{Osher06}. In this case, it is common to perform an alternative derivation of the gradient flow, which results in an expression similar to (\ref{E16}), with $\delta_\epsilon(\ls)$ replaced by $\|\nabla \ls\|$. The latter, in turn, could be approximated by 1, whenever $\ls$ is defined to be the signed distance function of its zero level-set~\cite{Osher06} (which is the case in the present study).

\section{Technical considerations}
\subsection{Regularization of curvature}
\begin{figure}[top]
\centering
\makebox[\textwidth]{\includegraphics[width=6in]{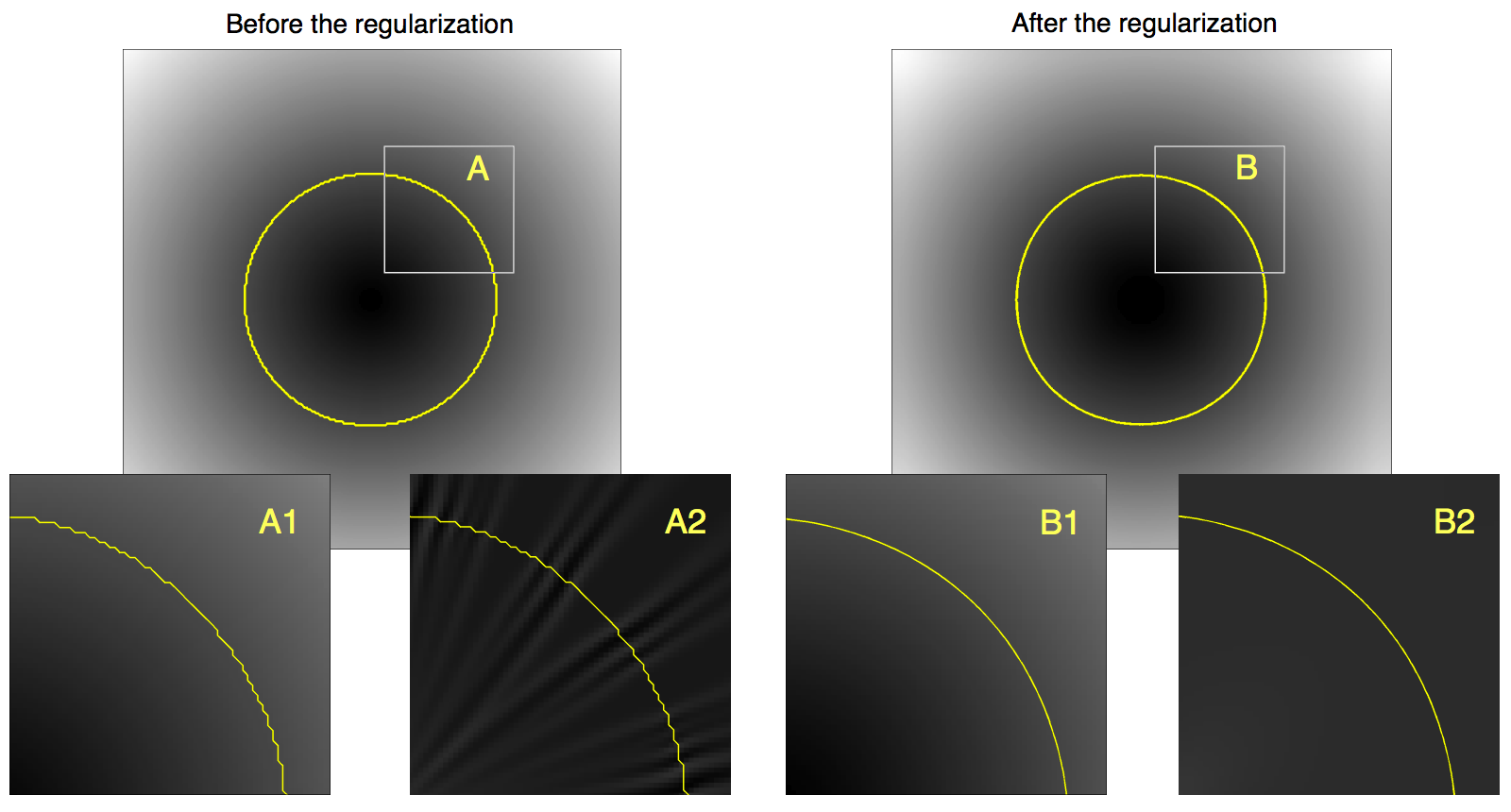}} 
\caption{(Subplot A) A level-set function and its corresponding zero level-set before the regularization; (Subplot A1) A zoomed fragment of the level-set function in Subplot A; (Subplot A2) The curvature of the level-sets in Subplot A1; (Subplot B) The regularized version of the level-set function in Subplot A; (Subplot B1) A zoomed fragment of the regularized level-set function in Subplot B; (Subplot B2) The curvature of the level-sets in Subplot B1.}  
\label{F2}
\end{figure}

As long as practical implementation of a level-set evolution is considered, it is standard to compute the curvature $\kappa(\xx)$ based on 
\begin{equation}\label{E17}
\kappa(\xx) = - {\rm div} \left\{ \frac{\nabla \varphi(\xx) }{\| \nabla \varphi(\xx) \|} \right\} 
= \frac{2\varphi_x\varphi_y\varphi_{xy}-\varphi_{xx}\varphi_y^2-\varphi_{yy}\varphi_x^2}{(\varphi_x^2+\varphi_y^2)^{3/2}},
\end{equation}
where the subscripts $x$ and $y$ denote partial differentiation along the corresponding directions. The partials, in turn, are normally computed by means of standard discretization schemes, whose numerical support is necessarily finite. In this case, both quantization and measurement noises can cause the discrete curvature to be a noisy, ``jagged" function~\cite{Macklin06}. Needless to say, such behavior of the curvature could bias the estimation of its corresponding pdf. This situation can be observed in Fig.~{\ref{F2}}, Subplot A of which shows a grayscale image of the level-set function $\ls$ that is defined to be the signed distance function of the circle depicted in yellow. A zoomed fragment of the circle and of its associated level-set function is shown in Subplot A1, while Subplot A2 depicts the corresponding values of $\kappa(\xx)$. Note that, theoretically, the curvature of a circle is a constant equal to the inverse of the circle's radius. However, due to the domain discretization, the ``discrete" circle appears to be a piecewise linear curve, and, as a result, the curvature of its corresponding $\ls$ varies considerably along the level-sets, as shown in Subplot A2.

One possible way to alleviate the problem of irregularity of $\kappa(\xx)$ is to increase the numerical support of discretization of the partial derivatives in (\ref{E17}). Unfortunately, in the case under consideration, this approach has not resulted in substantial improvements.  An alternative way to suppress the spurious irregularities of discrete $\kappa(\xx)$ is to precede its computation by a {\em regularization} of the corresponding $\ls$. In particular, we suggest to regularize $\ls$ via anisotropically diffusing the latter in the direction {\em tangent} to its level-sets. Such a diffusion can be readily performed using the approach of~\cite{Weickert99}, which prescribes to smooth $\ls$ through solving 
\begin{equation}\label{reg}
\frac{\partial\varphi(\xx,\tau)}{\partial \tau} = {\rm div} (D(\xx)\,\nabla\varphi(\xx,\tau)),
\end{equation}
with $D \in \mathbb{R}^{2\times 2}$ being a diffusivity matrix, and $\tau$ being an artificial (iteration) time similar to the one in (\ref{E17}). 

The requirement on the diffusion in (\ref{reg}) to propagate in the direction tangent to the level-sets of $\ls$ is controlled via a special definition of $D$. Specifically, $D$ can be defined to have the same eigenvectors $\vec{v}_1$ and $\vec{v}_2$ as those of a smoothed version of the structural tensor $\nabla \ls \nabla \ls^T$. Note that when the smoothing is negligible, the directions of $\vec{v}_1$ and $\nabla \ls$ coincide, implying $\vec{v}_1 \parallel \nabla\ls$ and $\vec{v}_2 \perp \nabla \ls$. (When the effect of smoothing cannot be ignored, the latter relationships will hold approximately.) Therefore, setting $D = \vec{v}_1 \vec{v}_1^T + \gamma \, \vec{v}_2 \vec{v}_2^T$, with $\gamma \ll 1$, guarantees the diffusion in (\ref{reg}) will smooth $\ls$ in the direction of its level-sets, while preserving the geometric shape of the zero level-set. In the experiments reported in this paper, the value of $\gamma$ was set to be equal to 0.1. Moreover, the diffusion in (\ref{reg}) was performed using the AOS-based procedure of~\cite{Weickert99}, with the number of iterations and the diffusion time-step being equal to $N=4$ and $\Delta \tau = 10$, respectively. 

The value of the above procedure can be appreciated via observing Subplot B of Fig.~\ref{F2}, which shows regularized versions of the circular curve and its associated $\ls$, which were previously discussed in relation to Subplot A of the same figure. One can see that the regularization based on (\ref{reg}) produces a much smoother level-set function, whose curvature better complies with its theoretically predicted value. 

The overall structure of the proposed method for segmentation of TRUS images is summarized in the pseudocode of Algorithm 1. The algorithm receives as its input a TRUS image $u(\xx)$, the target pdf $P_t(\zz)$ of a set of photometric features, the target pdf $C_t(\xi)$ of the curvature of prostate boundary, as well as an initial level-set function $\varphi_0(\xx)$. At the output, the algorithm returns a final level-set function $\varphi_t(\xx)$, which can be used to indicate the prostate region as $\left\{\xx \mid \varphi_t(\xx) \leq 0 \right\}$, or by means of its boundary $\left\{\xx \mid \varphi_t(\xx) = 0 \right\}$.   Note that at Step 10 of Algorithm 1 below, the operation denoted by ``AOS" refers to the procedure represented by equation (51) in~\cite{Weickert98}.  

\begin{algorithm}
\caption{Proposed segmentation procedure} 
\begin{algorithmic}[1]
\STATE {\textbf{given:}} $u(\xx)$, $P_t(\zz)$, $C_t(\xi)$, $\varphi_0(\xx) \equiv \varphi(\xx,t=0)$
\STATE {\textbf{preset:}} $\Delta \tau = 10$, $\alpha =0.5$, $\beta = 2.5$, $t=0$
\STATE {\textbf{compute:}} $\{I_k(\xx)\}_{k=1}^d$ and $g(\xx)$ using (\ref{E11}) (e.g. with $\lambda=3$) 
\WHILE {$\delta > 10^{-4}$}
\STATE {Diffuse $\varphi_t(\xx)$ using (\ref{reg}) to result in $\tilde{\varphi}_t(\xx)$}
\STATE {Compute $\kappa(\xx)$ using $\tilde{\varphi}_t(\xx)$ and equation (\ref{E17})}
\STATE {Compute $\left\{ p(z_k \mid \varphi_t(\xx)) \right\}_{k=1}^d$ using (\ref{E6}) and $C(\xi \mid \varphi_t(\xx))$ using (\ref{E13})} 
\STATE {Compute $V_B(\xx)$ using (\ref{A1-7}) and $V_C(\xx)$ using (\ref{B1-4})}
\STATE {$\varphi_{t+1}(\xx) \Leftarrow \varphi_t(\xx) + \Delta t \, \left( \alpha \, V_B(\xx) + \beta \, V_C(\xx) \right)$}
\STATE {$\varphi_{t+1}(\xx) \Leftarrow {\rm AOS}\left(\varphi_{t+1}(\xx), g(\xx),\Delta t\right)$}
\STATE {Redistance $\varphi_{t+1}(\xx)$ by fast marching}
\STATE {$\delta \Leftarrow \| \varphi_{t+1} - \varphi_{t} \|$}
\STATE {$t \Leftarrow t+1$}
\ENDWHILE
\RETURN {$\varphi_{t}(\xx)$}
\end{algorithmic}
\end{algorithm} 
 
\subsection{Feature selection}

\begin{figure}[top]
\centering
\makebox[\textwidth]{\includegraphics[width=6in]{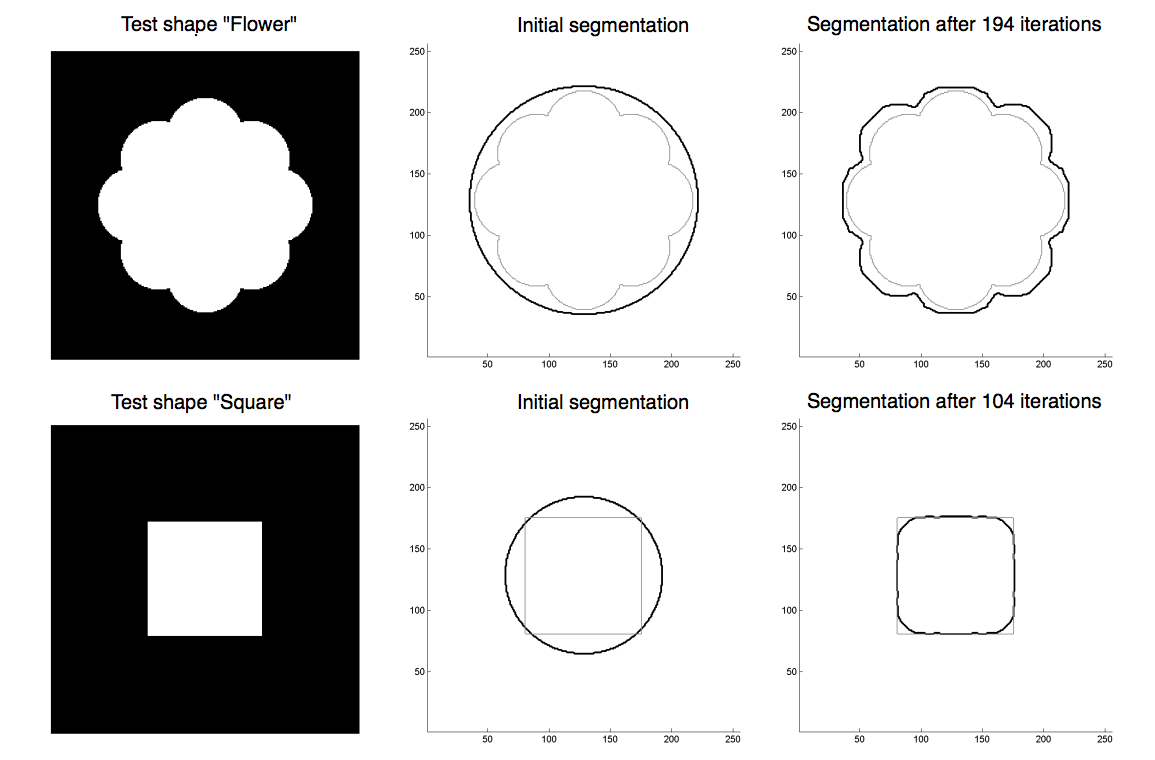}} 
\caption{(Upper row of subplots [left to right]) Test shape ``Flower", its initial and final segmentations based on $V_C(\xx)$ alone; (Lower row of subplots [left to right]) Test shape ``Square", its initial and final segmentations based on $V_C(\xx)$ alone.}
\label{F3}
\end{figure}

In the previous sections, we introduced a different way to incorporate prior shape information into the process of segmentation of TRUS images. It should be noted that, in general, the same distribution of $\kappa(\xx)$ can be shared by an infinite number of different shapes. For this reason, the proposed shape prior by itself cannot guarantee the segmentation to converge to a useful result. At the same time, the shape prior still contains essential information as demonstrated by the example of Fig.~\ref{F3}. The leftmost subplots of the figure show two test shapes, {\it viz.} ``flower" and ``square", while the middle and the rightmost subplots show their initial and final segmentation, respectively, using the shape velocity $V_C(\xx)$ alone. One can see that, despite the severe ambiguity of definition of the shape priors in the form of pdf's, the segmentations in Fig.~\ref{F3} are still capable of converging to a close vicinity of the true shapes.

The above examples are by no means generic, and therefore in order to ensure the convergence of segmentation in the general case of TRUS images, the use of image-related features is necessary. The arsenal of possible features is broad, with some typical examples including (yet not limited to) Gabor and wavelet transform coefficients~\cite{Porat89, Unser95}, multiresolution moments~\cite{Suhling04}, local fractal dimension~\cite{Chen89}, and many others~\cite{Laws80, Haralick73}. Another way to find a set of informative features could be to project a (large) set of arbitrary features onto a subspace spanned by their either independent or principal components~\cite{Hyvarinen01}.

Since the main contribution of this paper is the introduction of the pdf-based ``weak" shape priors, the experimental study of the next section does not provide an in-depth analysis of performance of the proposed segmentation method for different choices of image features. The only features of TRUS images used in the experiments are their gray-levels as well as the values of their SRAD filtered versions. Despite the relative simplicity of the above choice of features, the segmentation results appear to be very promising as demonstrated by examples below.

\section{Results}\label{Res}
\subsection{In silico experiments}
\begin{figure}[top]
\centering
\makebox[\textwidth]{\includegraphics[width=6in]{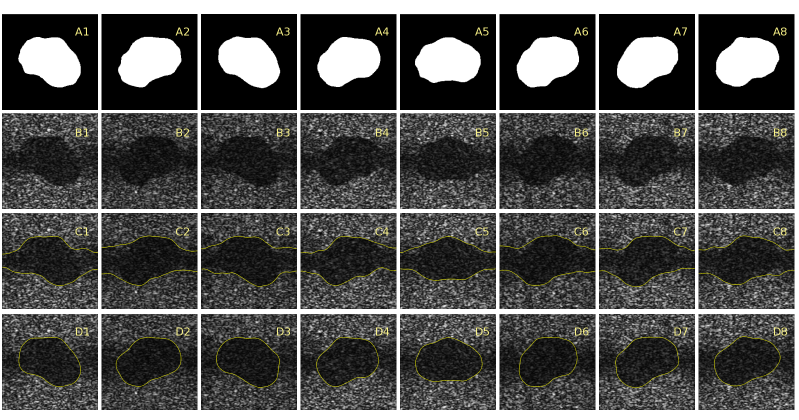}}
\caption{(First row of subplots) True segmentation profiles used for simulation of prostate images; (Second row of subplots) The corresponding prostate images (contrast 3:1) corrupted by shadowing artifacts; (Third row of subplots) Segmentation results obtained by Algorithm 1 with $\beta=0$ (i.e. without shape priors); (Fourth row of subplots) Segmentation results obtained by Algorithm 1 using the shape priors.}
\label{F5}
\end{figure}

In this section, the performance of the proposed algorithm is first validated through an initial stage of {\it in silico} experiments, followed by subsequent {\it in vivo} experiments. In particular, for the case of {\it in silico} validation, the RF-images of prostate were simulated by convolution of synthetic reflectivity functions with a point spread function (PSF). The latter was obtained experimentally by imaging a point-target ({\it viz.} a thin steel wire in a water tank) using a single-element, 3.5 MHz-transducer (Panametrics V383, Waltham, MA) for both transmission and reception. The lateral scanning of the target was carried out mechanically with a lateral resolution of 0.4 mm, and the acquired RF-lines were sampled at a rate of 25 MHz. The reflectivity functions, on the other hand, were generated as 2-D white Gaussian noise fields weighted by predefined amplitude profiles, which were designed to mimic the typical geometry of a prostate gland (see the first row of subplots in Fig.~\ref{F5} for illustrative examples). Moreover, to account for the variability of real-life data, the profiles were subjected to random (elastic) deformations.

\begin{figure}[top]
\centering
\makebox[\textwidth]{\includegraphics[width=6in]{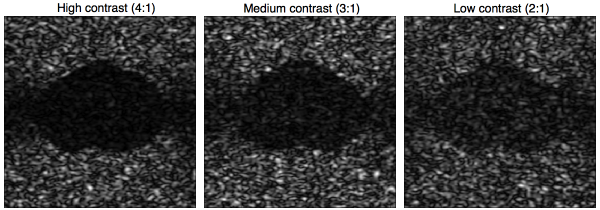}} 
\caption{Simulated prostate images of high (4:1), medium (3:1), and low (2:1) contrasts.}
\label{F4}
\end{figure}

The contrast of prostate regions was controlled by the values of the variance of simulated reflectivity functions inside and outside of the prostate boundaries. Three different ratios of the variances, {\it viz.} 4:1, 3:1, and 2:1, were used to test the performance of the segmentation under variable conditions. Moreover, to assess the robustness of the proposed method, the reflectivity functions were pre-multiplied by another mask to mimic the presence of shadowing artifacts as well as to cause the apparent boundaries of simulated prostates to be discontinuous. Some typical envelope images corresponding to the simulated RF-data are shown in Fig.~\ref{F4}, whose leftmost, middle, and rightmost subplots illustrate the high (4:1), medium (3:1), and low (2:1) contrasts, respectively. 

In the present study, 200 simulated prostate images were generated for each contrast category with half of them being used as training images and the rest used for algorithm validation. In particular, the training images together with their SRAD filtered versions were used to pre-compute the target pdf's $P_t(\zz)$ and $C_t(\xi)$. The test images, on the other hand, were used for segmentation, followed by a quantitative analysis of its performance. The latter was assessed in terms of the \emph{normalized mean squared error} (NMSE) criterion. Specifically, let $M$ be the matrix of the true binary template of a simulated prostate, and $\hat{M}$ be its estimate. Then, the NMSE can be defined as
\begin{equation}\label{NMSE}
{\rm NMSE} = \mathcal{E} \left\{ \frac{\| M - \hat{M} \|_F^2}{ \| M \|_F^2 }  \right\},
\end{equation}
where $\| \cdot \|_F$ denotes the Frobenius matrix norm, and $\mathcal{E}$ stands for the operator of expectation, which in the present case was approximated by a sample mean computed over the 100 test images. 

A subset of typical segmentation results are shown in Fig.~\ref{F5}. In particular, the upper row of subplots of the figure depicts the simulated templates of prostate shapes used in the simulation study, while the subplots in the second row show the corresponding envelope images. At the first stage, these images were segmented by Algorithm 1 without using the ``weak" shape priors (i.e. $\beta = 0$). The results of this segmentation are demonstrated by the third row of subplots in Fig.~\ref{F5}. One can see that the discontinuity of prostate boundaries along with the low contrast of the shadowed regions have caused the segmentation to converge to an erroneous result. To resolve this problem, the same images were segmented by Algorithm 1 with the inclusion of shape priors (i.e. $\beta = 2.5$). The resulting segmentation is shown in the bottom row of subplots of Fig.~\ref{F5}. One can see that in this case the segmentation is capable of converging to the true prostate shapes.

\begin{table}[here]
\caption{NMSE ($\pm$ one standard deviation) of prostate segmentation}
	\centering
	\noindent\makebox[\textwidth]{
	\begin{tabular}{l*{4}{c}r}
	\toprule
	\cmidrule(r){1-4}
	 & High Contrast (4:1) & Medium Contrast (3:1) & Low Contrast (2:1) \\
	\midrule
	NMSE (with priors) & 0.049$\pm$0.009 & 0.050$\pm$0.014 & 0.066$\pm$0.019 \\
	NMSE (without priors)  & 0.211$\pm$0.035  & 0.247$\pm$0.048 & 0.340$\pm$0.053 \\
	\bottomrule
	\end{tabular}
	\label{T1}}
\end{table}

The quantitative comparison results are summarized in Table~\ref{T1}, which shows the values of NMSE obtained with and without using the shape priors for different contrast levels. It can be clearly seen that the inclusion of the shape priors results in more than 5-fold decrease in the value of NMSE.

\subsection{In vivo experiments}
In addition to the {\it in silico} experiments, {\it in vivo} experiments were conducted next to further test the performance of the proposed algorithm.  For the latter case, the ultrasound images were obtained during clinical TRUS sessions using the Aloka 2000 ultrasound machine with a broadband 7 MHz linear transducer and a field of view of approximately 6 cm. Subsequently, the obtained set of clinical TRUS images were manually delineated by an expert radiologist. 

Apart from providing the ground truth for a comparative analysis, the manually delineated images were also used to learn the target pdf's $P_t(\zz)$ and $C_t(\xi)$. As before, the image features were defined to be the gray-level values of the TRUS images and their SRAD filtered versions, while the prostate morphology was described by the curvature of its boundary.  

\begin{figure}[top]
\centering
\makebox[\textwidth]{\includegraphics[width=5in]{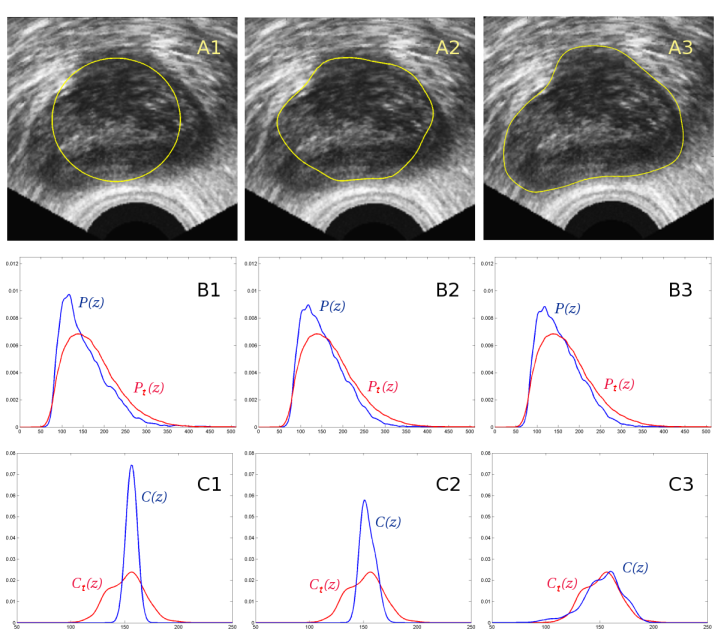}} 
\caption{(Subplots A1-A3) Initial, intermediate, and final segmentations of a prostate image; (Subplots B1-B3) The target (red) and empirical (blue) pdf's of the image gray-levels corresponding to the segmented regions shown in Subplots A1-A3, respectively; (Subplots C1-C3) The target (red) and empirical (blue) pdf's of the curvature of the active contours (yellow) shown in Subplots A1-A3, respectively.}
\label{F6}
\end{figure}

In the \textit{in vivo} experiments, it was found to be impossible to attain useful segmentation results without the use of shape priors. For this reason, only the prior-enhanced segmentation results are reported in this section. A typical segmentation process is demonstrated in Fig.~\ref{F6}. Starting at the left and transitioning to the right of the figure, each column represents the results of initial, intermediate, and final segmentation stages. In particular, the upper row of subplots depicts the convergence of the active contour from its initial to final configuration, while Subplots B1-B3 and Subplots C1-C3 compare the target (red) and empirical (blue) pdf's. Specifically, the pdf's corresponding to the intensity values of the TRUS images (as observed within the boundaries of the active contour) are plotted in Subplots B1-B3. At the same time, Subplots C1-C3 shows the evolution of the curvature pdf's. One can see that as the segmentation converges, the appearances of the target and empirical pdf's become progressively more similar. 

\begin{figure}[top]
\centering
\makebox[\textwidth]{\includegraphics[width=6in]{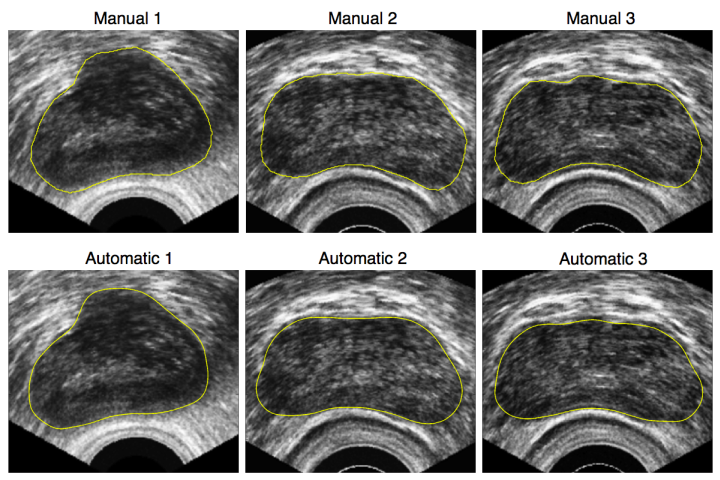}} 
\caption{Examples of TRUS image segmentation: (Upper row of subplots) Expert manual delineations; (Lower row of subplots) Results produced by Algorithm 1.}
\label{F7}
\end{figure}

Additional segmentation examples are demonstrated in Fig.~\ref{F7}. The shown images represent most of the principal challenges of TRUS image segmentation, as they suffer from severe shadowing artifacts and low contrast. Moreover, the boundaries of the shown prostates are discontinuous due to the aforementioned artifacts as well as the destructive influence of speckle noise. Despite the above  difficulties, the proposed method yields reliable and accurate segmentation, which closely matches the results of manual delineation. The NMSE of the \emph{in vivo} segmentation by means of Algorithm 1 was found to be equal to 0.081 $\pm$ 0.017.

\section{Discussion and Conclusions}
A new method of automatic segmentation of TRUS images of prostate was proposed in this paper. The method can be categorized as a learning-based technique, since it takes advantage of the availability of large sets of manually delineated TRUS data. The latter makes it possible to learn the statistical properties of discrimination features characterizing the prostate region before the actual segmentation is carried out. Although the list of possible segmentation features is rather long, only the intensities of TRUS images and their SRAD-filtered versions were used in the present study. While the resulting segmentation was found to be stable, robust, and accurate, we believe that further improvements are possible via a more meticulous selection of the discriminating features. 

The segmentation algorithm proposed in the current paper is based on the concept of distribution tracking, which provides an efficient and versatile framework for incorporation of an arbitrary number of statistical features in the process of image delineation. Moreover, one of the principal contributions of this research has been in introduction of a way in which the same set of ideas can be extended to tracking of the morphological features of prostates. This effort has resulted in developing the concept of ``weak" priors, which are ``encoded" in the form of pdf's of some geometric parameters of the expected prostate boundaries. In this case, the segmentation is forced to converge to the shapes, which have the same parameters distributed similarly to the target distributions. The main advantage of such priors consists in their property of being minimally restrictive, which suggests that the resulting estimates will have a negligible bias. On the other hand, the priors have been shown to be informative enough to render the segmentation stable and robust. 

The curvature of the prostate boundary was used as a shape descriptor to perform the tracking. It should be noted that, while being invariant under the group of Euclidean transformations, the curvature should not be used in the case when more complex (e.g. affine) deformations are expected. In this case, other geometric invariants should be considered. The definition of such invariants is in the focus of our current research. Among some other directions of our future work are an extension of the proposed method to 3-D scenarios as well as its verification via more extensive clinical experiments.

\section*{Appendix A}

To derive an expression for $\delta B(\varphi) \slash \delta \varphi$, we first note that the first variation of $p(z_k \mid \varphi(\xx))$ in (\ref{E7}) with respect to $\varphi(\xx)$ is equal to
\begin{equation}\label{A1-1}
\frac{\delta p(z_k \mid \varphi)}{\delta \varphi} = A^{-1} \delta(\varphi(\xx)) \Big( p(z_k \mid \varphi(\xx)) - K(z_k - I_k(\xx)) \Big),
\end{equation}
where $\delta(\cdot)$ is a Dirac delta function, and
\begin{equation}\label{A1-2}
A = \int_\Omega \mathcal{H}(-\varphi(\xx)) \, d\xx.
\end{equation}
Consequently, each of the $d$ coefficients $B_k(\varphi(\xx))$ in (\ref{E5}) has its first variation defined as
\begin{align}\label{A1-3}
\frac{\delta B_k(\varphi)}{\delta \varphi} &= \int_\mathbb{R} \frac{p_t(z_k)}{2 \sqrt{p_t(z_k) p(z_k \mid \ls)} } \, \frac{\delta p(z_k \mid \varphi)}{\delta \varphi} \, dz_k \notag \\
&= \delta(\ls) \, \frac{1}{2A} \left( B_k(\ls) - \int_\mathbb{R} \sqrt{\frac{p_t(z_k)}{p(z_k \mid \ls)}} \, K(z_k - I_k(\xx)) \, dz_k \right).
\end{align}
The above expression can be rewritten in a more concise form as
\begin{equation}\label{A1-4}
\frac{\delta B_k(\varphi)}{\delta \varphi} = \delta(\ls) \, \frac{1}{2A} \left( B_k(\ls) - \Big[r(z_k \mid \ls) \ast K(z_k) \Big]_{z_k = I_k(\xx)} \right),
\end{equation}
where $\ast$ stands for the operation of convolution and $r(z_k)$ is defined as
\begin{equation}\label{A1-5}
r(z_k \mid \ls) = \sqrt{\frac{p_t(z_k)}{p(z_k \mid \ls)}}.
\end{equation}
It should be pointed out that, in practical computations, the functions $r(z_k \mid \ls)$ and $K(z_k)$ are represented by their discrete values evaluated over some predefined subset of the real line (see Section~\ref{Res} for more details). In this case, the convolution in (\ref{A1-4}) can be efficiently computed by means of an FFT-based algorithm~\cite[Ch.8]{Oppenheim98}, followed by evaluating its result at the points $z_k=I_k(\xx)$ via, e.g., either linear or cubic interpolation.   

Finally, the results in (\ref{A1-4}) can be combined together to yield the first variation of $B(\ls)$ as given by
\begin{equation}\label{A1-6}
\frac{\delta B(\varphi)}{\delta \varphi} = \delta(\ls) \, V_B(\xx), 
\end{equation}
where
\begin{equation}\label{A1-7}
V_B(\xx) = \frac{1}{2A} \sum_{k=1}^d \alpha_k  \left( B_k(\ls) - \Big[ r(z_k \mid \ls) \ast K(z_k) \Big]_{z_k = I_k(\xx)} \right),
\end{equation}
with
\begin{equation}\label{A1-8}
\alpha_k = \prod_{i=1, i \neq k}^d B_i(\ls).
\end{equation}

\section*{Appendix B}
We derive the first variation of $B_\kappa(\ls)$ with respect to $\ls$ under the assumption that $\| \nabla \ls \| = 1$, which is valid as long as $\ls$ is initialized and sustained to be a signed-distance function, which is the case in the present study. Under this assumption, one can show that the first variation of $C(\xi \mid \ls)$ with respect to $\ls$ is given by 
\begin{equation}\label{B1-1}
\frac{\delta C(\xi\mid\varphi)}{\delta\varphi} = \frac{\Delta \left[\delta_\epsilon(\ls)K^\prime(\xi-\kappa(\xx))\right]+\delta_ \epsilon'(\ls)\left[K(\xi-\kappa(\xx))-C(\xi \mid \ls)\right]}{\int{\delta_ \epsilon (\ls)d\xx}}.
\end{equation}
Consequently, the first variation of $B_\kappa(\ls)$ in (\ref{E14}) is defined as
\begin{equation}\label{B1-2}
\frac{\delta B_\kappa(\varphi)}{\delta \varphi} = \frac{1}{2}\int \sqrt{\frac{C_t(\xi)}{C(\xi \mid \ls)}} \, \frac{\delta C(\xi \mid \varphi)}{\delta \varphi} \, d\xi.
\end{equation}
Let $L(\xi) = \left[ C_t(\xi) \slash C(\xi\mid\ls) \right]^{1/2}$ and $A = \int \delta_\epsilon(\xx) d\xx$. Then $\delta B_\kappa(\varphi) / \delta (\varphi)$ becomes
\begin{align}\label{B1-4}
V_C(\xx) \triangleq \frac{\delta B_\kappa(\varphi)}{\delta(\varphi)} &= \frac{1}{2} \int L(\xi) \, \frac{\delta C(\xi\mid\ls)}{\delta \ls} d\xi  = \notag \\
&=\frac{1}{2 \, A} \int L(\xi) \, \Delta\left[\delta_\epsilon(\ls) \, K^\prime(\xi-\kappa(\xx))\right]d\xi + \notag\\
&\hspace{1cm} + {{\delta_\epsilon'(\ls)}\over{2 \, A}} \int L(\xi) \, \left[K(\xi-\kappa(\xx))-C(\xi\mid\ls)\right]d\xi = \notag \\
&=\frac{1}{2 \, A} \Bigg[ \int L(\xi) \, \Delta\left[\delta_\epsilon(\ls) \, K^\prime(\xi-\kappa(\xx))\right]d\xi + \notag\\
& \hspace{1cm} + \delta_\epsilon'(\ls) \Big( \big[ L(\xi) \ast K(\xi) \big]_{\xi = \kappa(\xx)}  - B_\kappa(\ls)\Big) \Bigg],
\end{align}
where $\ast$ stands for the operation of convolution as before. As the next step, one could further expand the first (integral) term in (\ref{B1-4}). However, based on the results of numerical experiments, it was found that a more stable estimation of $\delta B_\kappa(\varphi) / \delta (\varphi)$ can be achieved by directly approximating the integral using the trapezoidal rule together with a standard discretization of the Laplacian. Note that this computation is numerically efficient, as it only needs to be performed over the subset $\{ \xx \mid \delta_\epsilon(\xx) \neq 0 \}$, which is small due to the finite support of $\delta_\epsilon(\cdot)$.

\bibliographystyle{IEEEtran}
\bibliography{IEEEabrv,references}

\end{document}